# On the use of first and second derivative approximations for biometric online signature recognition

Marcos Faundez-Zanuy [1[0000-0003-0605-1282]], Moises Diaz [2[0000-0003-3878-3867]]

[1] Tecnocampus, Universitat Pompeu Fabra, 08302 Mataró, Spain
[2] Universidad de Las Palmas de Gran Canaria, Spain
faundez@tecnocampus.cat

**Abstract.** This paper investigates the impact of different approximation methods in feature extraction for pattern recognition applications, specifically focused on delta and delta-delta parameters. Using MCYT330 online signature database, our experiments show that 11-point approximation outperforms 1-point approximation, resulting in a 1.4% improvement in identification rate, 36.8% reduction in random forgeries and 2.4% reduction in skilled forgeries

## 1      Online signature biometric recognition

Signatures are a widely utilized biometric modality in e-security systems based on the premise of "something you can do" [1][14]. Probably one of its main advantages is that the user can decide to change his signature when it is compromised. Unfortunately, this is not possible with most biometric traits such as face, speech, iris, etc. In addition, it has a long tradition of centuries as an authentication method in legal contracts, paintings, etc., and it can play an important role in health assessment too [2][14]. Several different handwritten tasks can be used [3].

Online biometric recognition can operate in two different ways:
a) Identification (1:N): The goal is to compare a given signature with the N models stored in a database of N users. Usually, the model that best fits the input signature indicates the identified user.
b) Verification (1:1): A user provides his signature. Then his claimed identity and the system tries to guess if he is a genuine or forger user [14]. Signature databases may contain two distinct types of forgeries, namely, random and skilled. In the skilled forgery type, the forger deliberately attempts to imitate the genuine signature. Conversely, in the random forgery type, the forger utilizes their own signature as a replacement for the genuine signature..

In this paper, we evaluate the relevance of delta and delta-delta parameter approximation in the feature extraction block.

The structure of the paper is as follows: Section 2 provides a comprehensive review of the relevant literature. Section 3 details the essential components of the recognition system, with a specific focus on feature extraction, including the employed databases, normalization techniques, and distance computation algorithm. Section 4 presents the empirical findings, while Section 5 summarizes the key outcomes and conclusions derived from this study.

## 2      Related works

The use of derivatives in parametrization, i.e., delta and delta-delta parameter, is a common technique in online signature verification. It can provide additional information about the signature and help to improve recognition accuracies.

There are several different methods that have been proposed for the use of derivatives in online signature verification, including the use of velocity, acceleration, and jerk features [14].

 In [15], the authors used delta coordinate differences between two consecutive points in x-y coordinates, achieving the lowest error rates with the delta parameter. This represented the velocity in x-y coordinates independently. In [16], the benefits of using the kth order derivative were described, where the authors only worked out the first and second derivative sequences of vectors. In the last signature verification competition [17], some participants used delta, such as DLVC-Lab or SIG team, who used the first and second-order derivatives. Similar features were selected by the authors of [18], who chose 12 features including the first- and second-order derivatives (delta and delta-delta, respectively), and applied a DTW algorithm for time sequence matching.

Typically, the derivatives were computed by subtracting two consecutive sampling points. However, [8] found good results using a second-order regression, which had the advantage of output features having the same length as the non-derivative ones. This regression formula was also followed in [19], where it was applied to robotic features estimated from x-y.

To the best of our knowledge, related works mostly use the first and sometimes the second derivative, but only few works have gone beyond a simple derivative operation with delta and delta-delta parameters.

## 3      Experimental setup

The general pattern recognition system depicted in Figure 1 is indeed suitable for signature recognition applications. It consists of four blocks, described next.

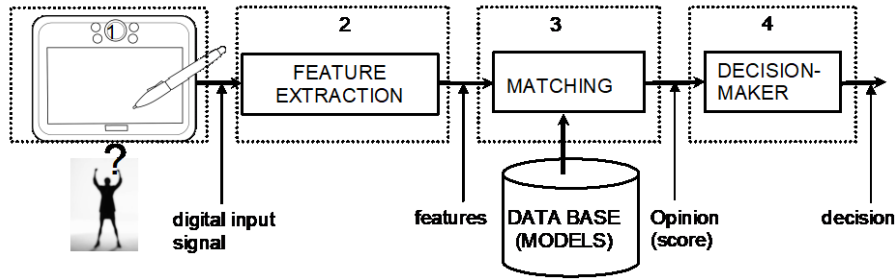

**Fig. 1.** General pattern recognition system

Block 1 (registered signature): digitizing tablet is used for online signature acquisition. In our case, we have used a well-known pre-existing database, which is MCYT [4], and is summarized in Section 3.1. The database is split into two parts: the training set, used for user model computation, and the testing set, used to provide experimental recognition rates.

Block 2 (Feature extraction): The digitizing tablet provides x, y, p, al, az information as well as a time stamp code. Knowing that the sample rate is 200 samples per second, the feature set can be extended as described in Section 3.2.

Block 3 (Matching): In this paper, we use the dynamic time warping (DTW) distance computation [5], which has been widely used as a successful recognition technique and is summarized in Section 3.3.

Block 4 (Decision maker): We used the DETWare V2.1 NIST toolbox [6-7] for identification and verification assessments.

### 3.1 MCYT Database

The MCYT signature database was acquired using a WACOM graphic tablet, with a sampling frequency of 100 Hz. Each sampled instance of the signature contains the following information:

1) Position along the x-axis, $x$ : [0–12 700], equivalent to 0–127 mm;
2) Position along the y-axis, $y$ : [0–9700], equivalent to 0–97 mm;
3) Pressure, $p$, applied by the pen: [0–1024];
4) Azimuth angle, $az$, of the pen with respect to the tablet (as shown in Figure 2): [0–3600], equivalent to 0–360 degrees;
5) Altitude angle, $al$, of the pen with respect to the tablet (as shown in Figure 2): [300–900], equivalent to 30–90 degrees.

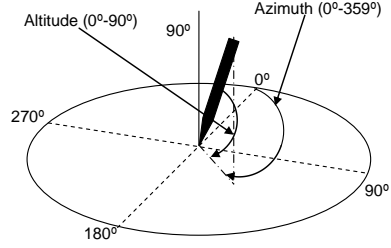

**Fig. 2.** Azimuth and altitude angles of the pen with respect to the plane of the WACOM tablet.

The MCYT signature database comprises a total of 330 users, each of whom contributed 25 genuine signatures and 25 skilled forgeries. The skilled forgeries were produced by the subsequent five target users, who were provided with static images of the genuine signature and instructed to imitate them (for at least ten times) in a natural manner without introducing any discontinuities or irregularities. This resulted in the acquisition of highly realistic skilled forgeries that captured the shape-based natural dynamics of genuine signatures. Specifically, user n produced a set of 5 genuine signature samples, followed by 5 skilled forgeries of client n-1, another set of 5 genuine signature samples, and 5 skilled forgeries of user n-2. This process was repeated for users n-3, n-4, and n-5. Thus, user n contributed a total of 25 samples of their own genuine signature and 25 skilled forgeries (5 final samples each from users n-1 to n-5). Similarly, for user n, 25 skilled forgeries were produced by users n+1 to n+5 using the same procedure.

### 3.2 Feature extraction and normalization

Each signature followed the next process:
1. The spatial coordinates ($x$, $y$) are normalized by the centroid or geometric center ($[\bar{x}, \bar{y}]$) (1) of each signature. Thus, the center of each signature is displaced to (0, 0) coordinate (2) as follows.

$$[\bar{x}, \bar{y}] = \frac{1}{L}\sum_{l=1}^{L}[x_l, y_l] \qquad (1)$$

$$x_l = x_l - \bar{x}, y_l = y_l - \bar{y} \qquad (2)$$

2. Feature set (*f*) is enlarged by working out the delta (first derivate) and delta-delta (second derivate) parameters. Thus, from the five feature set provided by the Intuos Wacom digitizing tablet $f = [x, y, p, az, al]$, we obtain an eight dimension feature set $f = [x, y, p, \dot{x}, \dot{y}, \dot{p}, \ddot{x}, \ddot{y}]$. This optimal feature set was obtained in [8], which discards the angles information: *az*, *al*.

Delta parameters ($\dot{f}_i$) and delta-delta ($\ddot{f}_i$) features are the first and second derivative, respectively, for $i \in [1, 5]$. The delta parameters are obtained in the following way [9]:

$$delta = \frac{\sum_{k=-M}^{M} k \cdot x[k]}{\sum_{k=-M}^{M} k^2} \qquad (3)$$

The delta value of a feature $x$ is an estimation of the local slope of a region that is centered on sample $x[k]$ and spans $M$ samples before and after the current sample. This approximation is obtained through the method of least squares. The size of the region is defined by the delta window length, which extends from $-M$ to $M$. The length of the delta window is determined by an odd integer that is equal to or greater than three. On the other hand, we can calculate a simple derivative based on one sample difference $\dot{f}_i[l] = f_i[l] - f_i[l-1]$, which can be obtained with de MATLAB *diff* function and is equivalent to the basic definition of derivative equation (4):

$$\dot{f}_i = \lim_{h \to 0} \frac{f_i(l+h) - f_i(l)}{h} \tag{4}$$

for h equal to one sample (5):

$$\dot{f}_i = \frac{f_i(l+1) - f_i(l)}{1} = f_i(l+1) - f_i(l) \tag{5}$$

In this case, we add one zero at the beginning of the sequence in order to obtain the same length $L$ for the feature set and its derivative.

Delta-delta is obtained by applying two consecutive times the delta equation.

We have used the audioDelta function from MATLAB in the audio toolbox. It is worthy to mention that delta and delta-delta parameters have a long tradition in speaker and speech recognition [9]. For this reason, they were probably included in the audio toolbox. However, their potential is probably underexploited in handwritten analysis.

3. Features are normalized through a z-score using the following equation, where each feature $f_i$ is subtracted by its mean and divided by its standard deviation.

$$\hat{f}_i = \frac{f_i - \bar{f}_i}{\mathrm{std}(f_i)} \tag{6}$$

Figure 3 shows a sample signature coordinates (x, y) before and after normalization. We can observe that the shape of the signature is preserved. In addition, this z-score normalization makes unnecessary the centroid normalization described before.

Figure 4 shows the dynamic information of the signature of one user. From top to bottom: x-axis values (*x*), y-axis (*y*), pressure (*p*), azimuth (*az*) and altitude (*al*). In this specific case, the signature length is 786 samples. We observe that the angles *az* and *al* are defined with less bits and they show small variations. Thus, lesser discriminative information can be expected from them.

In order to emphasize the importance of bilateral implementation of derivative rather than relying on a single difference (equation 5), we have generated a synthetic signal and applied the single difference approach and bilateral with 15 points over the signal $x$ generated in MATLAB with the next equation (7):

```
x=[1:200,200*ones(1,200)]+rand(1,400);
```
(7)

Equation 7 consists of a ramp of 200 samples followed by a flat region of 200 samples. Then, a Gaussian noise is added to each sample

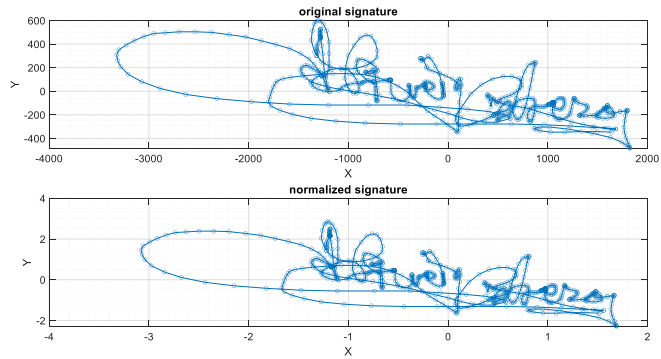

**Fig. 3.** Shape of a signature with raw coordinates and after the z-score normalization

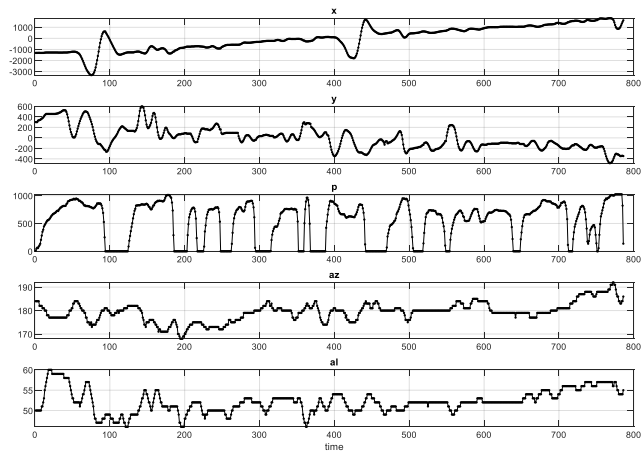

**Fig. 4.** Dynamic information of the signature of one user (x, y, p, az, al).

Figure 5 shows from top to bottom: the original signal x (equation 7), the normalized signal after applying equation 6 (xn), the normalization of the first derivative obtained with equation (4), and the first derivative with the audioDelta function and a window of 15 points. From this figure, we can clearly see the high impact of noise on the simple derivative approximation.

Worth to mention that fractional order derivatives are an active research field that provides good experimental results, especially in e-health based on handwritten tasks [10-12]. However, a deeper analysis must be done to discover if the improvement is due to the fractional order or the enlarged window analysis for derivative computation.

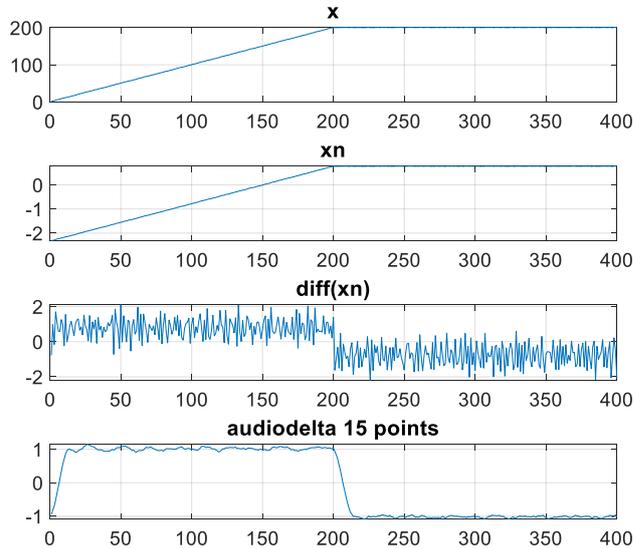

**Fig. 5.** Example of a synthetic signal x, normalized signal, first derivative approximated by two consecutive samples subtraction and first derivative with a 15-point window.

### 3.3 Dynamic time warping

In this study, feature matching with models is conducted by employing dynamic time warping (DTW), a well-known template matching algorithm that is highly effective in handling the random variations that arise due to intra-user variability [13]. DTW employs a dynamic programming approach to generate an elastic distance metric between two samples, regardless of any differences in their lengths. The algorithm is summarized in figure 6.

In addition, the DTW is normalized by the length of the first signature. Otherwise, those users with shorter signatures would tend to obtain smaller distance values. For this paper, we have used the standard DTW algorithm found in the signal processing toolbox in MATLAB.

Dynamic Time Warping (DTW) has several advantages over a machine learning approach for online signature recognition:
1. DTW is a non-parametric method, which means that it does not assume a specific functional form for the signature. This makes it more flexible than machine learning methods, which typically rely on a fixed model structure or distribution assumption.
2. DTW is robust to variations in speed and timing between signatures. In contrast, machine learning methods may require careful preprocessing or feature engineering to account for variations in timing or speed.

3. DTW is computationally efficient and does not require large amounts of training data. It can be applied to new signatures in real-time, making it suitable for online signature recognition applications. Machine learning methods, on the other hand, may require large amounts of training data and can be computationally expensive during both training and prediction phases.
4. DTW is interpretable and can provide insights into the similarity between signatures. Machine learning methods, on the other hand, may be more difficult to interpret and may not provide insights into the underlying similarity structure of the data. This is one of the main reasons to use DTW in this paper.

```
function dist = dtw_distance(sig1, sig2)
% Computes the Dynamic Time Warping distance between two online signatures.

% sig1: an array of shape (L1, 5) containing the first signature to be compared
% sig2: an array of shape (L2, 5) containing the second signature to be compared

% Initialize DTW matrix with zeros
dtw = zeros(length(sig1)+1, length(sig2)+1);

% Initializations
dtw(:, 1) = ∞;
dtw(1, :) = ∞;
dtw(1, 1) = 0;

% Compute DTW matrix
for i = 2:length(sig1)+1
   for j = 2:length(sig2)+1
      dist = norm(sig1(i-1,:) - sig2(j-1,:));
      dtw(i,j) = dist + min([dtw(i-1,j), dtw(i,j-1), dtw(i-1,j-1)]);
   end
end

% DTW distance (dist) is the bottom-right element of the DTW matrix
dist = dtw(end, end);
```

5. **Fig. 6.** DTW algorithm.

## 4  Experimental results

We have obtained a user model based on the first five training signatures. Then we performed three sets of experiments:

1. Identification: we have used five different testing signatures not used during the training process. Each testing signature was matched against all the models, which consisted of five different signatures. The minimum DTW distance from each testing signature to the five training signatures is selected (see Eq. 8). Then, the user that provides minimum distance is selected as the identified user (see Eq. 9), where $user \in [1, 330]$.

$$Distance_{user} = min(DTW(train1_{user}, test), \cdots, DTW(train5_{user}, test)) \quad (8)$$

$$user = min(Distance_1, \cdots, Distance_{user}, \cdots, Distance_{330}) \quad (9)$$

2. Verification with random forgeries: genuine user scores are obtained with DTW distances from a given user to its own model, while impostor scores are obtained with the DTW distances from a given user to the other's model. As the largest the DTW distance, the more different the signatures, we changed the sign of the DTW distance to convert from distance to score, as follows:

$$score_{user} = -Distance_{user} \quad (10)$$

The score set is the same as in the identification mode, but the addressed question is different. In verification, no identity is provided. Therefore, we modified the acceptance/rejection response, based on a decision threshold, which is adjusted by trial and error. This procedure implies a total amount of 330x5 genuine signatures, which produce the same amount of genuine distances, and 330x329x5 random forgeries (distances).

3. Verification with skilled forgeries. The process is analogous to the random forgeries. The genuine distances are the same of previous section. A new set of 330x25 distances are computed using the skilled forgeries.

We have repeated the whole set of experiments for a variable number of points used in the computation of the derivatives ranging from 1 to 15 (only odd numbers).

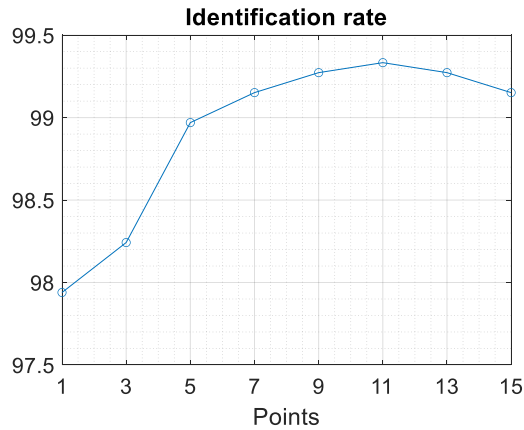

**Fig. 7.** Identification rate as function of the number of points used to work out the delta and delta-delta parameters.

Figure 7 shows the identification rates as a function of the number of points. In this case, we measured the accuracy of the identification task. Therefore, the higher the rate, the more precise the identification. As we can see, the highest value was achieved by using a window with 11 points.

On the other hand, Figure 8 shows the verification errors (minimum of the detection cost function or min(DCF) [6-7]) for random forgeries, and Figure 9 the min(DCF) for skilled forgeries.

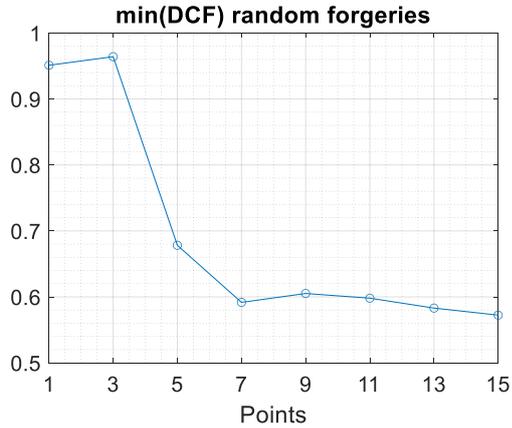

**Fig. 8.** Minimum of the detection cost function versus the number of points used to work out the delta and delta-delta parameters for random forgeries.

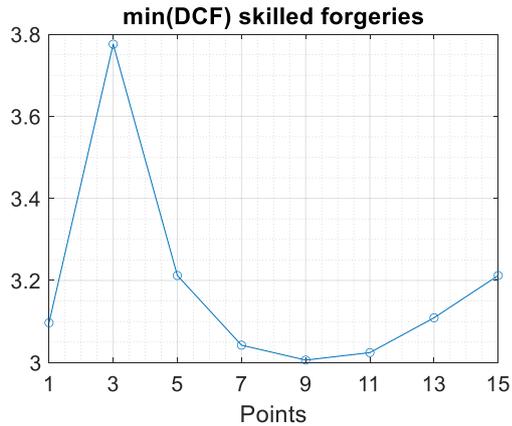

**Fig. 9.** Minimum of the detection cost function versus the number of points used to work out the delta and delta-delta parameters for skilled forgeries.

We observe that the best results were not obtained with the first delta in both verification cases. In the case of random forgery, an elbow effect was achieved in the plot. Although seven points change the rhythm of the plot, we can see a slight improvement when the number of points in the windows increases.

Observing the skilled forgeries results, we notice an "U" plot, which describes a minimum performance when nine points were used. In addition, a strange result was obtained when the delta and delta-delta were configured with a single point. It can confirm the extended use of one point in the derivates to extract features in different signature verification proposals in the literature.

Nevertheless, increasing the number of points in the delta and delta-delta negatively affects the execution time. For this reason, this is a trade-off problem between performance and execution optimization.

## 5      Conclusions

This article has shown the importance of configuring the delta (first derivative) and delta-delta (second derivative) to extract features in biometric signature recognition. Furthermore, we have analyzed the performance in several cases of delta and delta-delta for identification and verification problems, including random and skilled forgeries.

Looking at Figures 7, 6 and 7, we observe that the experimental implementation of the derivative significantly impacts the final performance. The simplest one based on a single point is not good enough. On the other hand, there is a limit on the number of points. After a certain value, the inclusion of more points in the computation window produces a drop in the results.

In view of previous results, we recommend using nine or eleven points for the computation of derivatives. Using eleven points we obtain a relative improvement of 1.4% in identification rate, 36.8% in random forgeries and 2.4% in skilled forgeries over the one point approximation.

**Acknowledgments**
This work has been supported by MINECO Spanish grant number PID2020-113242RB-I00, and PID2019-109099RB-C41.

## References

1. Faundez-Zanuy M., "Biometric security technology," in IEEE Aerospace and Electronic Systems Magazine, vol. 21, no. 6, pp. 15-26, June 2006, doi: 10.1109/MAES.2006.1662038.
2. Faundez-Zanuy, M., Fierrez, J., Ferrer, M.A. et al. Handwriting Biometrics: Applications and Future Trends in e-Security and e-Health. Cogn Comput 12, 940–953 (2020). https://doi.org/10.1007/s12559-020-09755-z
3. Faundez-Zanuy, M., Mekyska, J. & Impedovo, D. Online Handwriting, Signature and Touch Dynamics: Tasks and Potential Applications in the Field of Security and Health. Cogn Comput 13, 1406–1421 (2021). https://doi.org/10.1007/s12559-021-09938-2
4. Ortega-Garcia J., Fierrez-Aguilar J., Simon D., Gonzalez J., Faundez-Zanuy M., Espinosa-Duro V., Satue-Villar A., Hernaez I., Igarza J.-J., Vivaracho C., Escudero D., Moro Q.-I. "MCYT baseline corpus: a bimodal biometric database" Volume 150, Issue 6, December 2003, p. 395 – 401, DOI:  10.1049/ip-vis:20031078
5. Faundez-Zanuy M., "On-line signature recognition based on VQ-DTW" Pattern Recognition, Volume 40, Issue 3, 2007, Pages 981-992, ISSN 0031-3203, https://doi.org/10.1016/j.patcog.2006.06.007

6. Martin, A., Doddington, G., Kamm, T., Ordowski, M., Przybocki, M., "The det curve in assessment of detection performance". In: Proc. of the European Conf. on Speech Communication and Technology, pp. 1895–1898, 1997.
7. https://www.nist.gov/itl/iad/mig/tools
8. Fischer A., Diaz M., Plamondon R. and Ferrer M. A., "Robust score normalization for DTW-based on-line signature verification," 2015 13th International Conference on Document Analysis and Recognition (ICDAR), Tunis, Tunisia, 2015, pp. 241-245, doi: 10.1109/ICDAR.2015.7333760.
9. Rabiner, Lawrence R., and Ronald W. Schafer. Theory and Applications of Digital Speech Processing. Upper Saddle River, NJ: Pearson, 2010.
10. Mucha, J.; Mekyska, J.; Galaz, Z.; Faundez-Zanuy, M.; Lopez-de-Ipina, K.; Zvoncak, V.; Kiska, T.; Smekal, Z.; Brabenec, L.; Rektorova, I. Identification and Monitoring of Parkinson's Disease Dysgraphia Based on Fractional-Order Derivatives of Online Handwriting. Appl. Sci. 2018, 8, 2566. https://doi.org/10.3390/app8122566
11. Mucha J. et al., "Fractional Derivatives of Online Handwriting: A New Approach of Parkinsonic Dysgraphia Analysis," 2018 41st International Conference on Telecommunications and Signal Processing (TSP), Athens, Greece, 2018, pp. 1-4, doi: 10.1109/TSP.2018.8441293.
12. Mucha J. et al., "Analysis of Parkinson's Disease Dysgraphia Based on Optimized Fractional Order Derivative Features," 2019 27th European Signal Processing Conference (EUSIPCO), A Coruna, Spain, 2019, pp. 1-5, doi: 10.23919/EUSIPCO.2019.8903088.
13. Deller J. R., Proakis J. G., and Hansen J. H. L., "Dynamic Time Warping ," in Discrete-time processing of speech signals, New York: Macmillan Publishing Co., 1993
14. Diaz, M., Ferrer, M. A., Impedovo, D., Malik, M. I., Pirlo, G., & Plamondon, R. (2019). A perspective analysis of handwritten signature technology. Acm Computing Surveys (Csur), 51(6), 1-39
15. Kholmatov, A., & Yanikoglu, B. (2005). Identity authentication using improved online signature verification method. Pattern recognition letters, 26(15), 2400-2408.
16. Sae-Bae, N., & Memon, N. (2014). Online signature verification on mobile devices. IEEE transactions on information forensics and security, 9(6), 933-947.
17. Tolosana, R., Vera-Rodriguez, R., Gonzalez-Garcia, C., Fierrez, J., Rengifo, S., Morales, A. et al. (2021). ICDAR 2021 competition on on-line signature verification. In Document Analysis and Recognition–ICDAR 2021: 16th Interna-tional Conference, Lausanne, Switzerland, September 5–10, 2021, Proceedings, Part IV 16 (pp. 723-737). Springer International Publishing.
18. Jiang, J., Lai, S., Jin, L., and Zhu, Y. (2022). DsDTW: Local Representation Learning With Deep soft-DTW for Dynamic Signature Verification. IEEE Transactions on Information Forensics and Security, 17, 2198-2212.
19. Diaz, M., Ferrer, M. A., and Quintana, J. J. (2018, August). Robotic arm motion for verifying signatures. In 2018 16th International conference on frontiers in handwriting recognition (ICFHR) (pp. 157-162). IEEE.